\documentclass[letterpaper]{article} %
\usepackage[]{aaai24}  %
\usepackage{times}  %
\usepackage{helvet}  %
\usepackage{courier}  %
\usepackage[hyphens]{url}  %
\usepackage{graphicx} %
\usepackage{natbib}  %
\usepackage{caption} %
\usepackage{algorithm}
\usepackage{algorithmic}

\usepackage{array}
\newcolumntype{T}{>{\tiny}l} %

\usepackage{multirow}
\usepackage[table,xcdraw]{xcolor}

\usepackage{subcaption}

\usepackage{newfloat}
\usepackage{listings}
\DeclareCaptionStyle{ruled}{labelfont=normalfont,labelsep=colon,strut=off} %
\floatstyle{ruled}
\newfloat{listing}{tb}{lst}{}
\floatname{listing}{Listing}
\title{Evaluating Fairness in Self-supervised and Supervised Models for Sequential Data}
\author{
    Sofia Yfantidou \textsuperscript{\rm 2}\thanks{Work done at Nokia Bell Labs}\footnote{Corresponding author: syfantid@csd.auth.gr}
    Dimitris Spathis \textsuperscript{\rm 1}
    Marios Constantinides \textsuperscript{\rm 1}
    Athena Vakali \textsuperscript{\rm 2}
    Daniele Quercia \textsuperscript{\rm 1}
    Fahim Kawsar \textsuperscript{\rm 1}
}
\affiliations{

     \textsuperscript{\rm 1}Nokia Bell Labs Cambridge, United Kingdom
    
    \textsuperscript{\rm 2}Aristotle University of Thessaloniki, Greece

}

\usepackage{xcolor,color,soul,bm}

\newcommand{\rev}[1]{\textcolor[rgb]{0.00,0.00,0.00}{#1}}

\usepackage{bibentry}
\begin{document}

\maketitle

\begin{abstract}
Self-supervised learning (SSL) has become the de facto training paradigm of large models where pre-training is followed by supervised fine-tuning using domain-specific data and labels.
Hypothesizing that SSL models would learn more generic, hence less biased, representations, this study explores the impact of pre-training and fine-tuning strategies on fairness (i.e., performing equally on different demographic breakdowns).
Motivated by human-centric applications on real-world timeseries data, we interpret inductive biases on the model, layer, and metric levels by systematically comparing SSL models to their supervised counterparts. Our findings demonstrate that SSL has the capacity to achieve performance on par with supervised methods while significantly enhancing fairness---exhibiting up to a 27\% increase in fairness with a mere 1\% loss in performance through self-supervision. Ultimately, this work underscores SSL's potential in human-centric computing, particularly high-stakes, data-scarce application domains like healthcare.
\end{abstract}

\section{Introduction\label{introduction}}
The availability of extensive unlabeled, real-world timeseries data captured through various multimodal sensors presents significant opportunities. As a case in point, in the healthcare domain, leveraging this wealth of information can uncover intricate physiological and behavioral patterns at an unprecedented scale, offering novel insights into personalized and proactive healthcare \cite{perez2021wearables}.
However, the absence of ground-truth labels poses a significant obstacle to progress in supervised machine learning for timeseries tasks \cite{spathis2022breaking}. 
Self-supervision has proven its performance robustness and beyond state-of-the-art capabilities mainly in the areas of computer vision (CV) \cite{chen2020simple} and natural language processing (NLP) \cite{devlin2018bert} research. Inspired by such research, which leverages massive amounts of unlabeled data, many research communities that deal with timeseries data have swiftly recognized the potential of self-supervised learning (SSL). SSL exploitation is promising to explore extensive unlabeled data, effectively complementing small, labeled domain datasets \cite{tang2020exploring}.

Due to the recency of SSL adoption, in the cases of timeseries, performance metrics, such as accuracy and AUC-ROC score, are typically used as the prime evaluation criteria. Yet, a performance-centric evaluation approach can result in discriminatory or unjust impacts when comparing across different demographics. For instance, \citet{kamulegeya2019using} found that neural network algorithms trained to perform skin lesion classification showed approximately half the original diagnostic accuracy on black patients. At the same time, people of color are consistently misclassified by health sensors such as oximeters as they were validated on predominantly white populations~\cite{sjoding2020racial}. 

While SSL models may avoid such pitfalls due to their pre-training without (potentially) biased human annotations \cite{ramapuram2021evaluating}, comprehensive efforts to compare the fairness of supervised and SSL models are lacking. Note that fairness assessments typically assess accuracy disparities among diverse protected attributes, namely sensitive personal characteristics, such as race or gender, that are legally safeguarded from discrimination. In this study, we aim to address this gap by investigating how fine-tuning SSL models affects fairness in terms of outcomes and latent representations when compared to supervised alternatives. In detail, we make the following contributions:
\begin{itemize}
    \item For the first time, we look into fine-grained differences in the layer, model, and metric level between supervised and SSL models. Moving away from conventional performance-centric assessments, our approach adopts a human-centric perspective, integrating fairness metrics into our methodology to evaluate how these differences affect model outcomes. 
    
    \item We train and conduct a systematic comparison of a large number ($>$30) models with various levels of supervision and fine-tuning on multi-year health records timeseries data. We show that SSL performs on par with supervised models, but more notably, we observe up to a 27\% increase in fairness, accompanied by only a 1\% loss in performance for certain fine-tuning strategies.
    \item We compare the learned representations using the latent similarity between supervised and SSL models, which reveals discrepancies in latent representations across different demographic groups.
\end{itemize}

\section{Related Work}
While SSL methods such as SimCLR \cite{chen2020simple}, BYOL \cite{grill2020bootstrap}, and CPC \cite{oord2018representation}, have seen widespread use in CV \cite{kolesnikov2019revisiting}, NLP \cite{lan2019albert}, and audio \cite{saeed2021contrastive},
their application in unlabeled, real-world timeseries data has also been successful \cite{haresamudram2022assessing,spathis2021self}.

Existing works have extensively benchmarked SSL algorithms across domains, primarily focusing on performance metrics. However, limited attention has been given to evaluating fairness in self-supervised methods, particularly for multimodal, timeseries data. For example, in the healthcare setting, representation learning has been applied to patient monitoring \cite{yeche2021neighborhood}, Atrial Fibrillation detection \cite{tonekaboni2021unsupervised}, mortality or decompensation prediction \cite{harutyunyan2019multitask}, maternal and fetal stress detection \cite{sarkar2021detection}, and human-activity recognition \cite{tang2020exploring}, among others. However, the above works focus on performance-centric assessments.
Yet, the absence of biased annotations in SSL does not ensure fairness, necessitating evaluations beyond accuracy. As a matter of fact, studies in CV and NLP have demonstrated similarities with supervised alternatives in intermediate representations, emphasizing the need for fairness considerations \cite{grigg2021self,chung2019unsupervised}.

Fairness evaluations in SSL have been more prevalent in CV \cite{ramapuram2021evaluating} and NLP, including advances in generative models \cite{sheng2019woman}. Discussions on SSL's impact on fairness include training without prior data curation and the effects of fine-tuning \cite{goyal2022vision,ramapuram2021evaluating}. \rev{However, the examination of fine-tuning considerations in SSL primarily refers to updating Batch Normalization statistics and training residual skip connections, rather than assessing the impact of supervision level.}
In human-centric timeseries data, fairness evaluations have seen limited exploration, mainly in a supervised setting. For instance, for in-hospital mortality using the MIMIC-III or MIMIC-IV dataset \cite{meng2022interpretability,roosli2022peeking} or keyword spotting for on-device ML \cite{10.1145/3591867}.
All in all, such efforts are still in their early stages \cite{yfantidou2023beyond}, and an exploration of fairness in SSL for timeseries data is still lacking.

This paper aims to address the research gap by assessing SSL approaches in real-world human-centric data, considering both performance and fairness aspects.
While there are existing works addressing performance, fairness, or learned representations individually in SSL (across different domains), evidence that connects all these three aspects, particularly in human-centric timeseries data, is still lacking.

\section{Method}
\label{sec:method}
\begin{figure}[htb!]
  \centering
  \includegraphics[trim={2cm 0 2.2cm 0cm},clip,width=.9\linewidth]{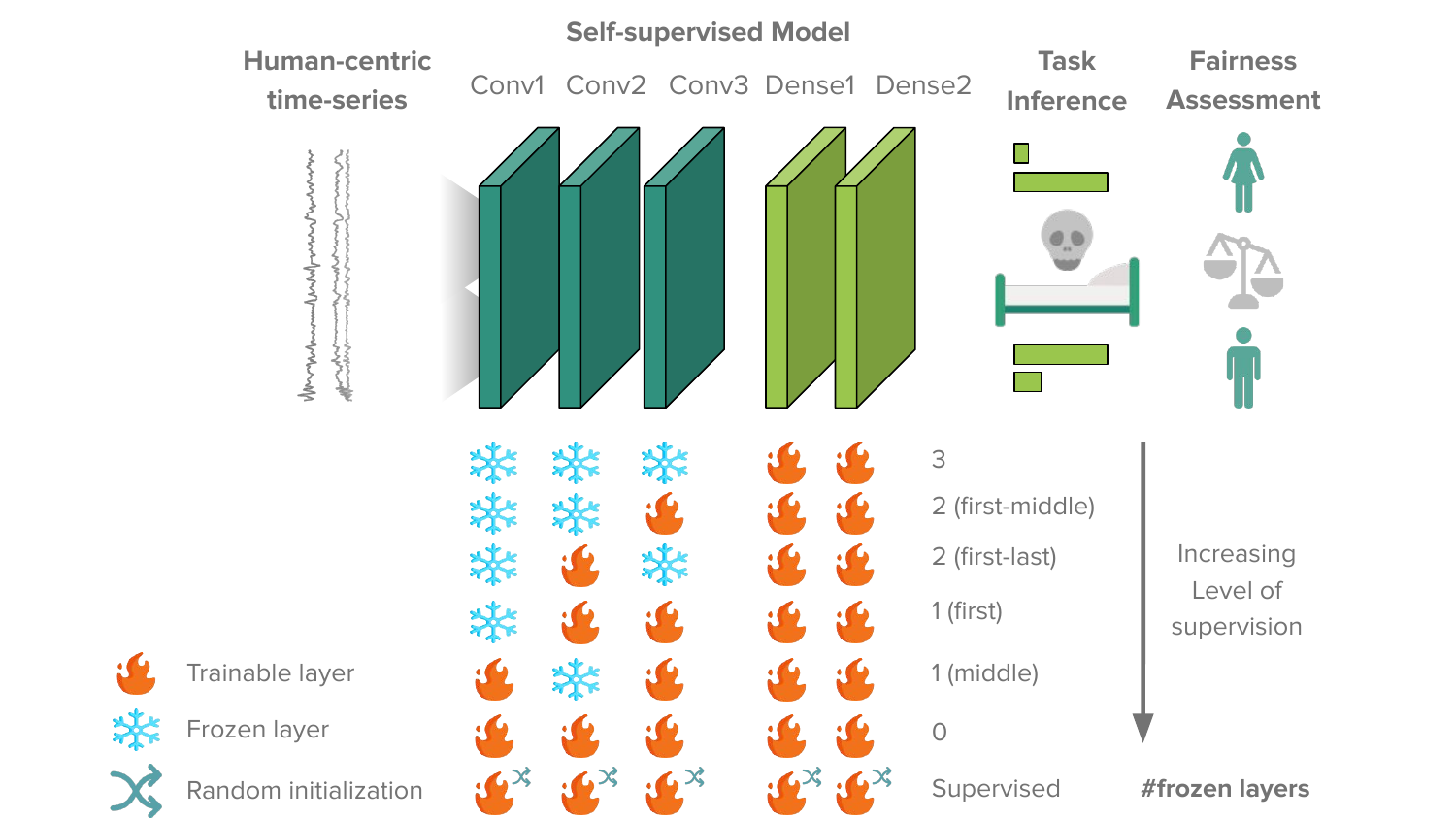}
  \caption{\textbf{Fine-tuning strategies for SSL workflows.} We experiment with various levels of supervision to optimize both performance and fairness.}
  \label{fig:modelsummaries}
  \vspace{-0.25cm}
\end{figure}

Having established that prior SSL literature has not adequately studied the impact of the level of supervision on fairness for human-centric timeseries, we conduct systematic experiments to investigate the extent to which fine-tuning strategies in SSL (Figure~\ref{fig:modelsummaries}) affect fairness and learned representations. Based on prior work on using contrastive learning in timeseries \cite{tang2020exploring}, we aim to answer the following research questions within the context of health timeseries tasks: How do supervision and fine-tuning levels affect fairness? What are the differences in learned representations between supervised learning and SSL?

\noindent\textbf{Models.} 
We use a SimCLR \cite{chen2020simple} variant by \citet{tang2020exploring}, adapted for (health) timeseries data. Our design mirrors SimCLR's key components: a) a \textit{stochastic data augmentation} module employing scaling and signal inversion; b) a 3-layer Convolutional Neural Network (CNN) as the \textit{base encoder}; c) a \textit{projection head} using a 2-layer Multi-layer Perceptron (MLP) for mapping representations; and d) a contrastive loss function, namely a normalized temperature-scaled cross-entropy loss (NT-Xent) \cite{sohn2016improved,chen2020simple}, as per \citet{tang2020exploring}. We define a similar architecture for the supervised baseline, replacing the contrastive loss with categorical cross-entropy.

\smallskip
\noindent\textbf{Fine-tuning Setup.} To assess the effect of self-supervision on fairness outcomes and representations, we employ a gradual freezing strategy, balancing the impact of the pre-trained encoder and the downstream labels (i.e., by increasing the  \#trainable parameters). We start by freezing all three base encoder layers and fine-tuning only the projection head. Then, gradually, we unfreeze layers one by one until both the encoder layers and projection head are fully trainable (similar to a fully supervised approach). We experiment with different freezing combinations, as seen in Figure~\ref{fig:modelsummaries}.\hfill
\smallskip 

\noindent\textbf{Representation Similarity.}
For assessing the impact of supervision on learned representations, we utilize recent advances in comparing neural representations. Specifically, we utilize Centered Kernel Alignment (CKA) as a similarity index, addressing challenges regarding the distributed nature, potential misalignment, and high dimensionality of representations \cite{kornblith2019similarity}. Note that we also condition CKA on protected attributes to identify differences in representation similarity across demographic groups.

\section{Evaluation}
\label{sec:evaluation}

\noindent\textbf{Dataset.} We employ a real-world timeseries dataset, the MIMIC-III Clinical Database \cite{johnson2016mimic}, in our evaluation. It contains over 31 million clinical events corresponding to 17 clinical variables.
These events cover 42.2K intensive care unit (ICU) stays of 33.8K unique subjects who were treated between 2001 and 2012. Our task involves the prediction of in-hospital mortality from observations recorded within 48 hours of an ICU admission---a primary outcome of interest in acute care. Following the benchmark by \citet{harutyunyan2019multitask}, we proceed with a total of 17,903 ICU stays and 3,236 subjects, forming 21.1K windows, each with 48 timestamps, 76 channels, and no overlap.

\smallskip
\noindent\textbf{Protected attributes.} To investigate fairness, we assess test-time performance for protected attribute groups, considering the documented variability in health timeseries based on physiological attributes \cite{spathis2021self}.
The MIMIC dataset contains a multitude of protected attributes relevant to the in-hospital mortality task: gender, age, ethnicity, religion, language, and insurance type (a proxy for socioeconomic status). For instance, prior work with MIMIC has revealed disparate treatment in prescribing mechanical ventilation among patient groups across ethnicity, gender, and age \cite{meng2022interpretability}, and voiced general fairness concerns for Black and publicly insured ICU patients \cite{roosli2022peeking}, indicating the importance of considering a diverse range of protected attributes. 
\smallskip

\noindent\textbf{Evaluation and fairness metrics.} To assess the performance of our models, we employ the AUC-ROC metric and calculate 95\% confidence intervals (CI). Consistent with the benchmark by \citet{harutyunyan2019multitask}, our focus on per-instance accuracy for mortality prediction leads to calculating overall performance as the micro-average across all predictions, irrespective of the patient. Regarding fairness, in line with prior work in high-stakes scenarios \cite{purdy2023pursuit}, we adopt the Error Rate Ratio (ERR) metric, namely the ratio of error rates for unprivileged and privileged groups computed as $\frac{ER_{D=unprivileged}}{ER_{D=privileged}}$, where $ER=\frac{FP+FN}{P+N}$ \cite{bellamy2019ai}. Values between 0.8 and 1.25 are considered fair, with 1.0 indicating parity.

\rev{We base our decision to use Error Rate Ratio on the worldviews it represents. From a technical perspective, Error Rate Ratio requires that the false positive and false negative rates be the same across protected user groups. Contextually, this means that given an outcome for a patient's mortality prediction, Error Rate Ratio requires that the probability of a corresponding incorrect prediction label be the same across all protected groups. More practically, this means Error Rate Ratio discourages using the protected attribute as a proxy for the outcome, promoting predictions independent of demographic factors \cite{hardt2016equality}. Conceptually, Error Rate Ratio aims to address population inequity, striving for risk predictions that avoid disproportionate harm to any group, irrespective of underlying risk distributions \cite{green2020false}.}
In the context of healthcare, \rev{balancing risk predictions is crucial}, given the potential consequences of false positives and false negatives and the ethical implications of \rev{risk} prediction disparities \cite{burt2017burden}. 

\smallskip
\noindent\textbf{Training Setup.}
Following \citet{tang2020exploring}'s recommended architecture for contrastive learning on health timeseries, our model comprises a base encoder featuring three temporal (1D) convolutional layers with kernel sizes of 24, 16, 8, and 32, 64, 96 filters, ReLU activation, a dropout rate of 0.1, and a concluding global maximum pooling layer. For pre-training, a projection head with three fully-connected layers (256, 128, and 50 units) is utilized, while the fine-tuned evaluation incorporates a classification head with two fully-connected layers (128 and 2 units). Pre-training employs the SGD optimizer with cosine decay of the learning rate over 200 epochs and a batch size of 128. Linear evaluation involves training for 100 epochs with the Adadelta optimizer and a learning rate of 0.03. 
All hyperparameters have been fine-tuned through manual adjustment.

\begin{table}
\centering
\caption{\textbf{Comparison of supervised and SSL models, conditioned on protected attributes.} The $\Delta$ columns illustrate differences between a specific segment and the general population, where yellow indicates disadvantaged and green indicates advantaged 
segments. The most disadvantaged segment is underlined, and the most advantaged one is bolded.}
\renewcommand{\arraystretch}{1.25}
\label{tab:aucroctable}
\resizebox{\columnwidth}{!}{%
\begin{tabular}{lllTllTll}
\hline
 &
   &
  \multicolumn{5}{c}{\textbf{Models}} &
   &
   \\ \hline
\textbf{Prot. Attribute} &
  \textbf{Segments} &
  \multicolumn{2}{l}{\textit{Supervised}} &
  $\mathit{\Delta_{gen.-segm.}}$ &
  \multicolumn{2}{l}{\textit{SSL (1 $\bullet\circ\bullet$)}} &
  $\mathit{\Delta_{gen.-segm.}}$ &
  $\mathbf{\Delta_{Sup.-SSL}}$ \\ \hline
Gen. Population &
  \textbf{} &
  \textit{0.839} &
  \textit{(0.82-0.86)} &
  \textit{} &
  \textit{0.829} &
  \textit{(0.81-0.85)} &
  \textit{} &
  \textit{} \\ \hline
 &
  $<$65 &
  0.863 &
  (0.83-0.89) &
  \cellcolor[HTML]{EAF8E5}0.024 &
  0.845 &
  (0.8-0.88) &
  \multicolumn{1}{l|}{\cellcolor[HTML]{F2FBEF}0.016} &
  \cellcolor[HTML]{FAFEF8}0.008 \\ \cline{2-9} 
\multirow{-2}{*}{Age} &
  $\ge$65 &
  0.822 &
  (0.8-0.85) &
  \cellcolor[HTML]{FEF9CA}-0.017 &
  0.82 &
  (0.79-0.85) &
  \multicolumn{1}{l|}{\cellcolor[HTML]{FEFBE0}-0.009} &
  \cellcolor[HTML]{FAFEF8}0.008 \\ \hline
 &
  White &
  0.839 &
  (0.82-0.86) &
  \cellcolor[HTML]{FEFEF9}0,000 &
  0.831 &
  (0.81-0.86) &
  \multicolumn{1}{l|}{\cellcolor[HTML]{FFFFFF}0.002} &
  \cellcolor[HTML]{FEFDF3}-0.002 \\ \cline{2-9} 
 &
  \underline{Black} &
  \underline{0.762} &
  (0.65-0.85) &
  \cellcolor[HTML]{FDE725}-0.077 &
  \underline{0.759} &
  (0.63-0.85) &
  \multicolumn{1}{l|}{\cellcolor[HTML]{FDE938}-0.070} &
  \cellcolor[HTML]{FBFEFA}0.007 \\ \cline{2-9} 
 &
  Asian &
  0.811 &
  (0.68-0.92) &
  \cellcolor[HTML]{FEF5AC}-0.028 &
  0.813 &
  (0.63-0.94) &
  \multicolumn{1}{l|}{\cellcolor[HTML]{FEF9CD}-0.016} &
  \cellcolor[HTML]{F6FCF4}0.012 \\ \cline{2-9} 
\multirow{-4}{*}{Ethnicity} &
  Hispanic &
  0.955 &
  (0.9-0.99) &
  \cellcolor[HTML]{8FDA77}0.116 &
  0.937 &
  (0.86-0.98) &
  \multicolumn{1}{l|}{\cellcolor[HTML]{97DC81}0.108} &
  \cellcolor[HTML]{FAFEF8}0.008 \\ \hline
 &
  Male &
  0.855 &
  (0.83-0.88) &
  \cellcolor[HTML]{F2FBEF}0,016 &
  0.843 &
  (0.81-0.87) &
  \multicolumn{1}{l|}{\cellcolor[HTML]{F4FCF1}0.014} &
  \cellcolor[HTML]{FFFFFF}0.002 \\ \cline{2-9} 
\multirow{-2}{*}{Gender} &
  Female &
  0.821 &
  (0.79-0.85) &
  \cellcolor[HTML]{FEF8C7}-0.018 &
  0.812 &
  (0.78-0.84) &
  \multicolumn{1}{l|}{\cellcolor[HTML]{FEF9CA}-0.017} &
  \cellcolor[HTML]{FEFEFC}0.001 \\ \hline
 &
  Medicare &
  0.825 &
  (0.8-0.85) &
  \cellcolor[HTML]{FEFAD2}-0.014 &
  0.819 &
  (0.79-0.84) &
  \multicolumn{1}{l|}{\cellcolor[HTML]{FEFBDD}-0.010} &
  \cellcolor[HTML]{FEFFFD}0.004 \\ \cline{2-9} 
 &
  Private &
  0.868 &
  (0.83-0.9) &
  \cellcolor[HTML]{E5F7DF}0.029 &
  0.856 &
  (0.81-0.9) &
  \multicolumn{1}{l|}{\cellcolor[HTML]{E7F7E2}0.027} &
  \cellcolor[HTML]{FFFFFF}0.002 \\ \cline{2-9} 
 &
  Medicaid &
  0.788 &
  (0.67-0.88) &
  \cellcolor[HTML]{FDEE6C}-0.051 &
  0.786 &
  (0.68-0.87) &
  \multicolumn{1}{l|}{\cellcolor[HTML]{FDF182}-0.043} &
  \cellcolor[HTML]{FAFEF8}0.008 \\ \cline{2-9} 
 &
  Government &
  0.885 &
  (0.77-0.99) &
  \cellcolor[HTML]{D4F1CB}0.046 &
  0.895 &
  (0.8-0.98) &
  \multicolumn{1}{l|}{\cellcolor[HTML]{C0EAB3}0.066} &
  \cellcolor[HTML]{FEF8C2}-0.020 \\ \cline{2-9} 
\multirow{-5}{*}{Insurance} &
  \textbf{Self Pay} &
  \textbf{0.983} &
  (0.93-1.0) &
  \cellcolor[HTML]{73D055}0.144 &
  \textbf{0.944} &
  (0.84-1.0) &
  \multicolumn{1}{l|}{\cellcolor[HTML]{90DA78}0.115} &
  \cellcolor[HTML]{E5F7DF}0.029 \\ \hline
 &
  English &
  0.839 &
  (0.81-0.87) &
  \cellcolor[HTML]{FEFEF9}0.000 &
  0.831 &
  (0.79-0.86) &
  \multicolumn{1}{l|}{\cellcolor[HTML]{FFFFFF}0.002} &
  \cellcolor[HTML]{FEFDF3}-0.002 \\ \cline{2-9} 
\multirow{-2}{*}{Language} &
  Other &
  0.831 &
  (0.8-0.86) &
  \cellcolor[HTML]{FEFBE3}-0.008 &
  0.82 &
  (0.79-0.84) &
  \multicolumn{1}{l|}{\cellcolor[HTML]{FEFBE0}-0.009} &
  \cellcolor[HTML]{FEFEF6}-0.001 \\ \hline
\end{tabular}%
}
\vspace{-0.4cm}
\end{table}

\begin{figure}[b]
    \centering
    \includegraphics[width=.9\linewidth]{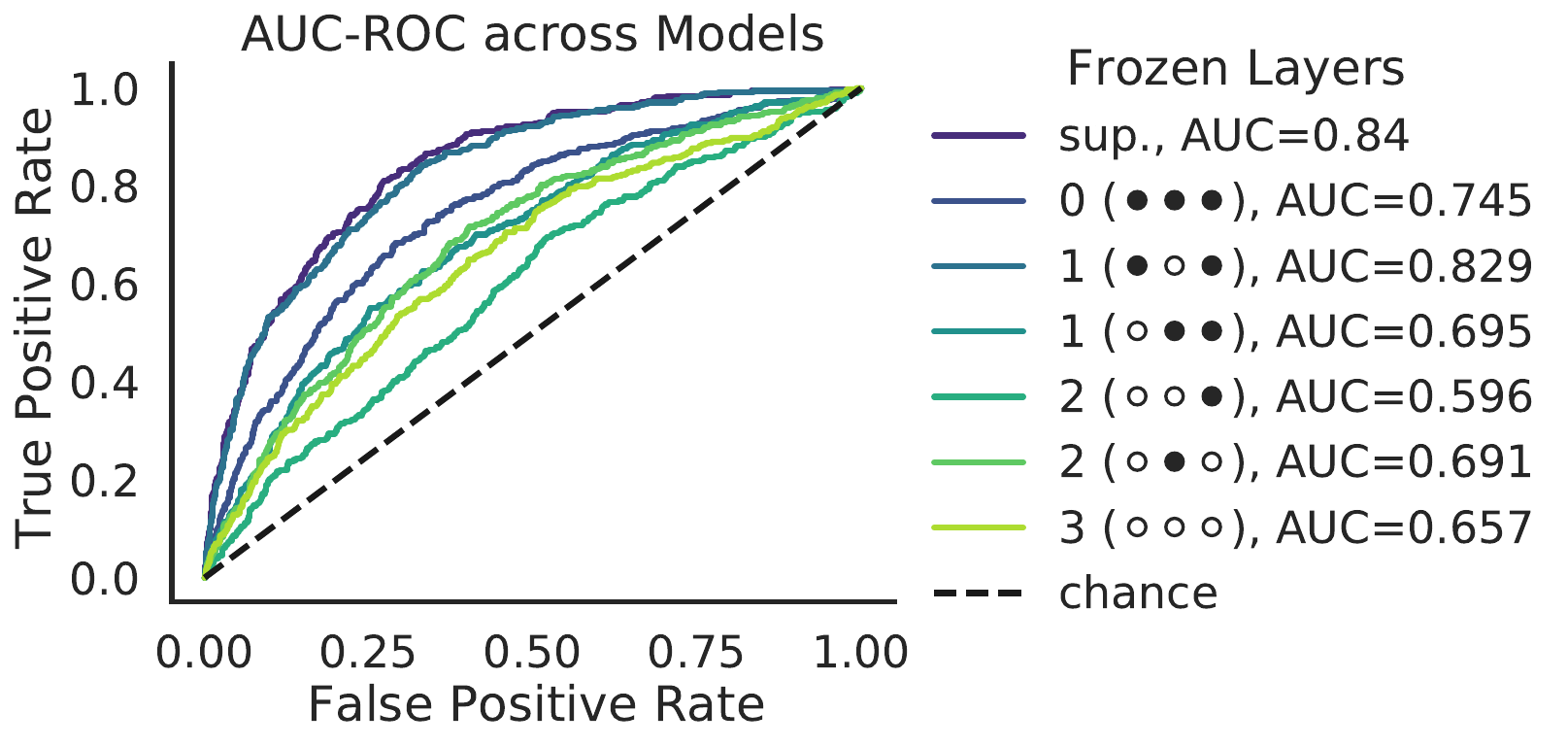}
    \caption{\textbf{AUC-ROC curves across models.} Depending on the fine-tuning strategy, SSL models achieve comparable performance to their supervised counterparts.}
    \label{fig:AUC-ROC}
    \vspace{-0.3cm}
\end{figure} %
\section{Results \& Discussion}\label{sec:results}
This section discusses the impact of supervision and fine-tuning in SSL on performance, fairness, and representations.

\subsection{Impact of supervision on performance}\label{model-evaluation}
Figure~\ref{fig:AUC-ROC} presents the ROC Curve and the AUC-ROC scores for both supervised and SSL models with various levels of fine-tuning. We notice that for the MIMIC dataset, the fully supervised model performs the best in terms of AUC-ROC with a score of 0.839 (CI 0.82-0.86). It is closely followed by the SSL model with a single frozen layer (middle) during fine-tuning, i.e., 1 ($\bullet\circ\bullet$) with an AUC-ROC score of 0.829 (CI 0.81-0.85), a mere 1\% loss in overall performance.

Table~\ref{tab:aucroctable} presents AUC-ROC scores per protected group for the supervised and the best-performing self-supervised model. We notice that the models do not perform equitably for all segments. Specifically, there exists a considerable performance gap for black patients, registering a deviation of nearly $-8\%$ in AUC-ROC, followed by Medicaid-insured patients with deviations exceeding $-5\%$. Conversely, patients with self-insurance show the best performance with deviations up to $+14\%$, trailed by Hispanics with deviations over $+11\%$. 
These findings align with previous studies involving supervised models for MIMIC-III mortality prediction. Notably, Medicaid patients consistently receive inferior predictions despite sharing comparable mortality rates with privately and self-insured patients. Similarly, black patients consistently underperform compared to white patients, even in the presence of lower mortality rates in the dataset. On the other hand, Hispanic patients exhibit elevated performance attributable to their significantly lower mortality rates compared to other demographic groups \cite{roosli2022peeking}.

Overall performance discrepancies seem similar between the two models if we focus on the $\Delta_{Supervised-SSL}$, with a slight fairness benefit for the SSL model. Yet, performance metrics are not always the best indicator for fairness. Even if a model performs well on average, it might exhibit significant differences in error rates across different groups. For instance, false positives or false negatives may disproportionately affect certain demographic groups, leading to unfair outcomes--a prospect we explore in the following section.

\begin{figure}[htb!]
    \centering
    \includegraphics[width=.9\linewidth]{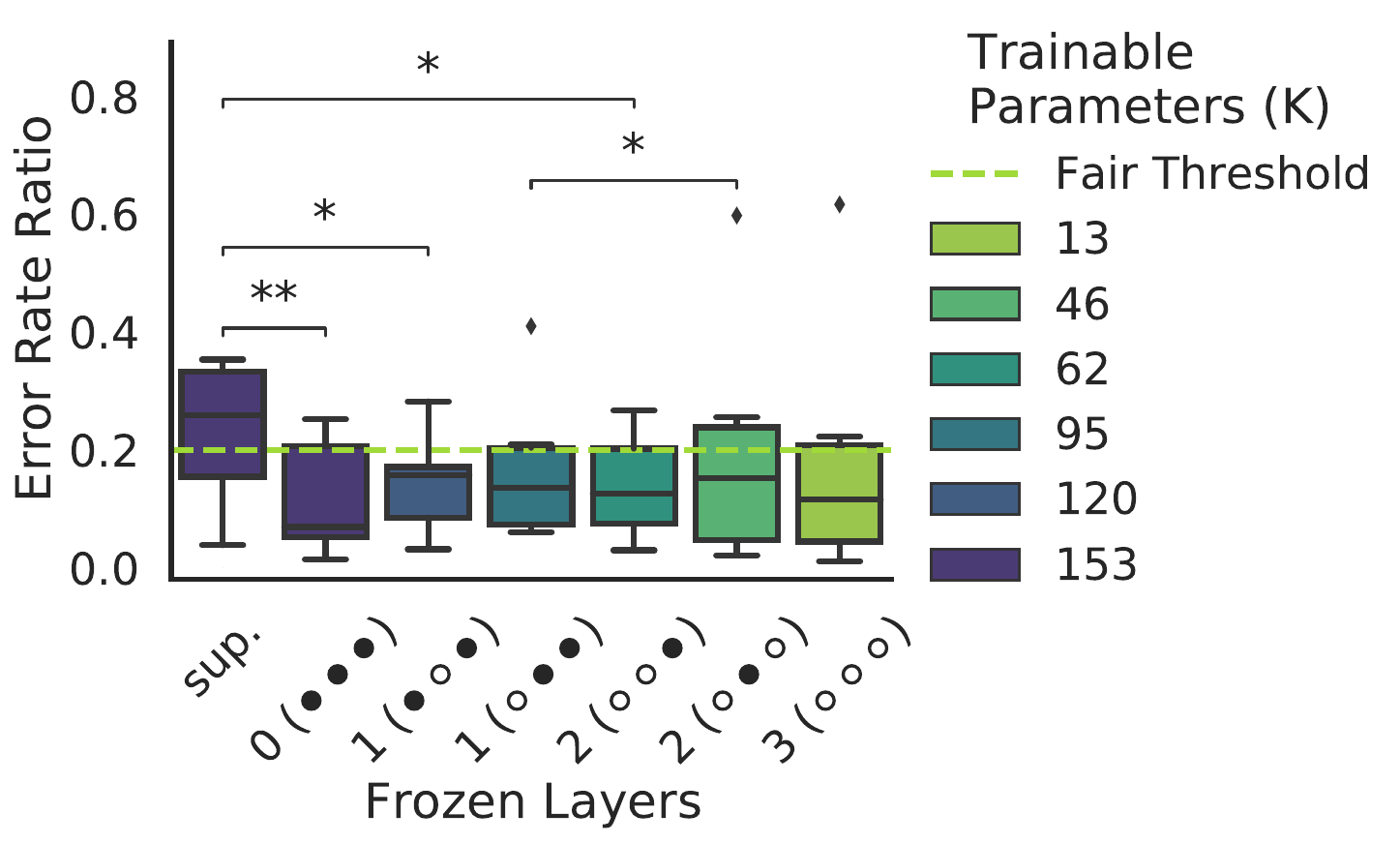}
    \caption{\textbf{Deviation from parity in error rate ratio across models.} While the supervised model has superior performance, it has a greater deviation from parity (dashed line) in terms of error rate ratio compared to the best-performing SSL model (i.e., 1 $\bullet\circ\bullet$).}
    \label{fig:mimic_fairness}
    \vspace{-0.5cm}
\end{figure}
\subsection{Impact of supervision and fine-tuning on fairness}\label{fairness-evaluation}
Figure~\ref{fig:mimic_fairness} visualizes the deviation of each model's error rate ratio --our fairness metric of choice-- from parity, i.e., $ERR=1.0$ (dashed line). Deviations greater than 0.2 indicate bias towards a protected attribute, irrespective of priviledge. While the supervised model has superior performance, it has a statistically significant ($p<0.05$) greater deviation from parity in terms of error rate ratio compared to the best-performing SSL model (i.e., 1 $\bullet\circ\bullet$). Specifically, the supervised model has a 0.125 deviation from parity, while the SSL one a 0.095--a 27\% decrease.
More importantly, all SSL models lie mostly within the acceptable limits, opposite to the supervised alternative. Note that prior work in other domains supports that model fine-tuning has an important impact on fairness \cite{ramapuram2021evaluating, rani2023self}. Indeed, our findings illustrate this point for timeseries data, too, as there exist differencies in error rate ratios between SSL models, even statistically significant ones (e.g.,  1 $\circ\bullet\bullet$ and 2 $\circ\bullet\circ$).

\begin{figure}
\begin{subfigure}[]{0.35\linewidth}
    \includegraphics[trim={0 0.4cm 3cm 0},clip,width=\linewidth]{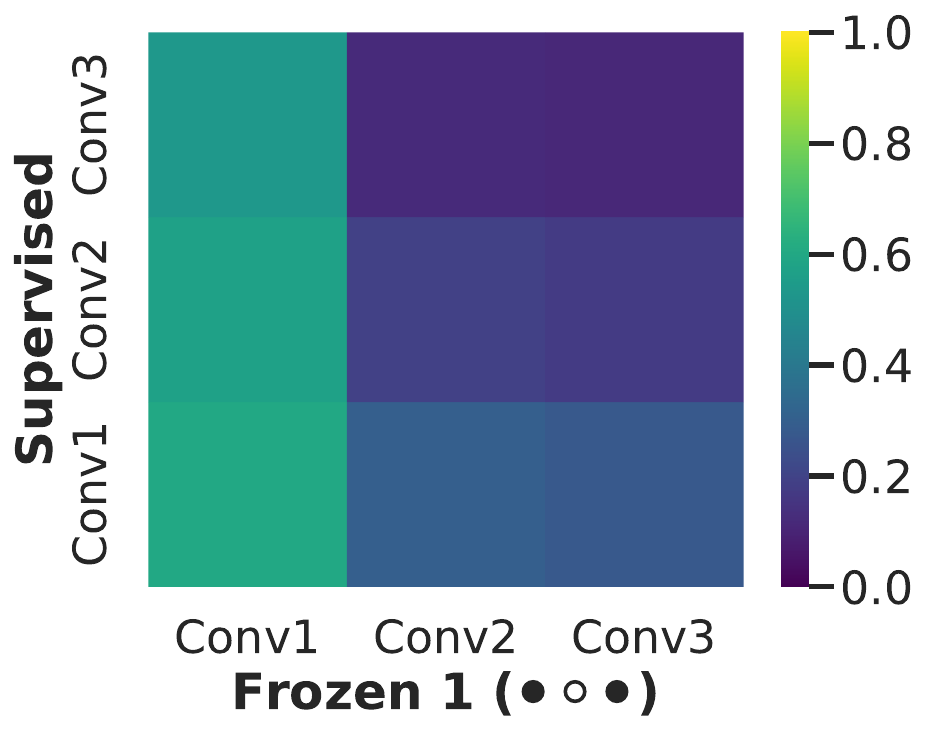}
    \caption{Random}
\end{subfigure}
\begin{subfigure}[]{0.27\linewidth}
    \includegraphics[trim={3cm 0.4cm 3cm 0},clip,width=\linewidth]{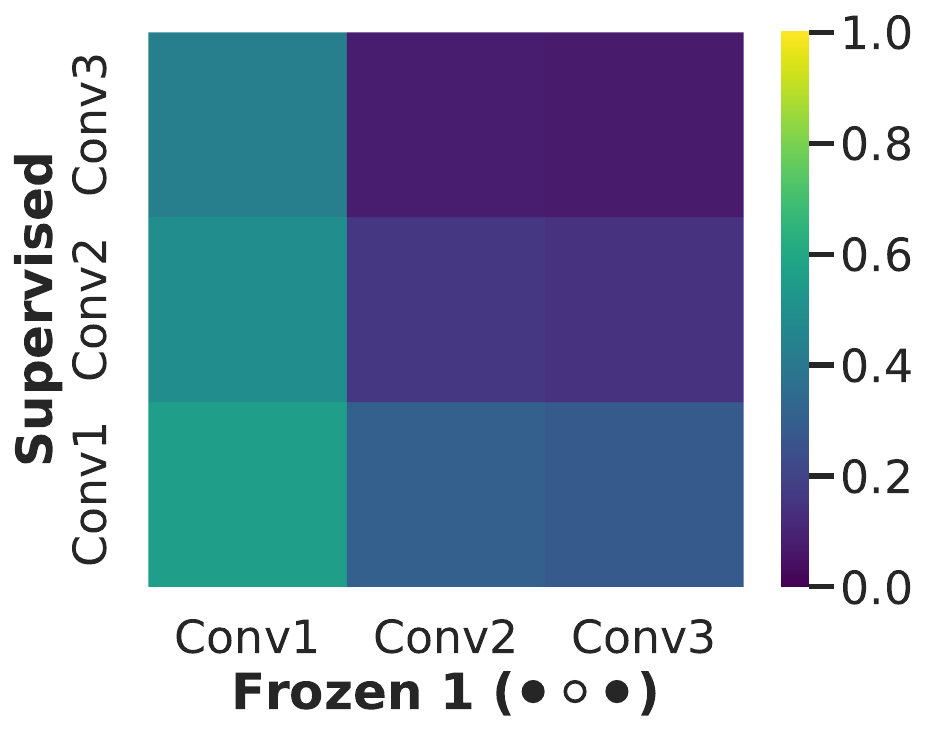}
    \caption{Female}
\end{subfigure} %
\begin{subfigure}[]{0.35\linewidth}
    \includegraphics[trim={3cm 0.4cm 0 0},clip,width=\linewidth]{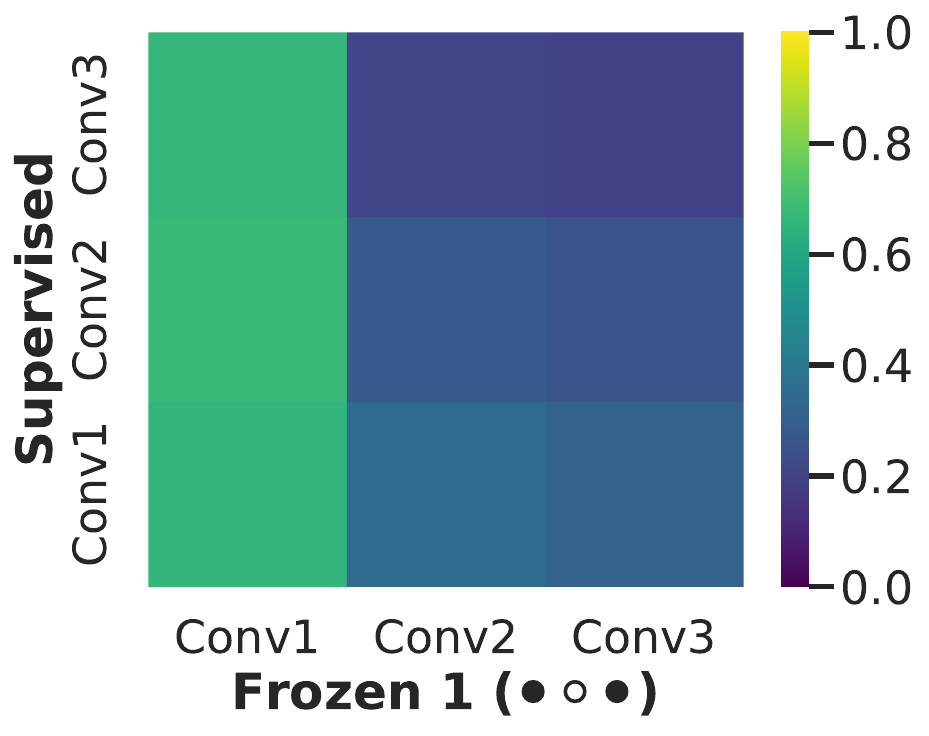}
    \caption{Male}
\end{subfigure} %
\caption{\textbf{Representation similarity through CKA conditioned on gender between the supervised and the (best) SSL models.} The random data subset is balanced to represent both genders equally. The similarity of the SSL and supervised models is higher for male than female users.}
\label{ckaconditionedgender}
\vspace{-0.4cm}
\end{figure}
\subsection{Interplay of representation similarity and fairness}\label{representation-evaluation}
We compare representation similarity between the supervised and the best-performing SSL model, i.e., 1 $\bullet\circ\bullet$, across protected attributes through CKA. Our findings regarding the impact of supervision on representation learning for timeseries data align with prior work on CV. Specifically, \citet{grigg2021self} illustrate how self-supervised and supervised methods learn similar visual representations through dissimilar means and that the learned representations diverge rapidly in the final few layers. Indeed, as illustrated in Figure~\ref{ckaconditionedgender}, the initial layer representations are similar, indicating a shared set of primitives. However, we notice discrepancies at the level of similarity for certain protected attributes. Taking gender as a case in point, we notice greater similarity in learned representations for males than for females, which could partially explain the superiority of the SSL models during the fairness assessment.

\section{Conclusions}
Our study reveals that SSL models can match fully supervised models' performance while demonstrating enhanced fairness, depending on fine-tuning. Specifically, moderate supervision during fine-tuning leads to self-supervised models exhibiting greater fairness across protected attributes. The impact on learned representations, measured through CKA, suggests that self-supervised models capture different data aspects, contributing to observed fairness variations.
While our evaluation focuses on the large multi-year MIMIC dataset, future work should replicate these findings in other tasks beyond critical care. Nevertheless, our study underscores SSL's potential benefits in harnessing scarce labeled data in real-world, timeseries-rich application domains, such as healthcare. Ultimately, it stresses the need for assessing fairness alongside performance metrics when considering machine learning models for real-world applications.

\section{Acknowledgments}
The authors affiliated with Aristotle University would like to acknowledge funding from the European Union’s Horizon 2020 research and innovation programme under the Marie Skłodowska-Curie grant agreement No 813162. The content of this paper reflects only the authors' view and the Agency and the Commission are not responsible for any use that may be made of the information it contains. 
Results presented in this work have been produced using the Nokia Bell Labs and the Aristotle University of Thessaloniki Compute Infrastructure and Resources.

\bibliography{aaai24}

\end{document}